\documentclass[11pt]{article}

\usepackage[final]{acl}

\usepackage{times}
\usepackage{latexsym}
\usepackage{multirow}
\usepackage{enumitem}
\usepackage{amsfonts}
\usepackage{amsmath}
\usepackage{booktabs}
\usepackage{multirow}
\usepackage{subcaption}
\usepackage{booktabs}
\usepackage{multirow}
\usepackage[table]{xcolor}
\usepackage{bm}
\usepackage{soul}
\usepackage{multirow}
\usepackage[T1]{fontenc}
\usepackage[normalem]{ulem}
\usepackage[utf8]{inputenc}
\usepackage{svg}

\usepackage{inconsolata}
\usepackage{enumitem}
\setlist{topsep=0pt, partopsep=0pt, itemsep=0pt, parsep=0pt}
\usepackage{graphicx}
\makeatletter
\g@addto@macro\normalsize{%
  \setlength\abovedisplayskip{6pt plus 2pt minus 2pt}%
  \setlength\belowdisplayskip{6pt plus 2pt minus 2pt}%
  \setlength\abovedisplayshortskip{4pt plus 1pt minus 1pt}%
  \setlength\belowdisplayshortskip{4pt plus 1pt minus 1pt}%
}
\makeatother
\title{RASPER: Reward-Aligned Summarization of Clinical Notes\\for EHR Outcome Prediction}

\author{ Arya Hadizadeh Moghaddam, Mohsen Nayebi Kerdabadi \\
\textbf{ Chen Chen, Dongjie Wang, Zijun Yao}\thanks{Corresponding author.} \\
Electrical Engineering and Computer Science, University of Kansas, USA \\
\texttt{\{a.hadizadehm, mohsen.nayebi, chenchen, wangdongjie, zyao\}@ku.edu}
}

\begin{document}
\maketitle
\begin{abstract}
Unstructured discharge notes in Electronic Health Records (EHRs) often carry signal complementary to structured medical codes, holding patient-specific evidence that standardized cohort-level codes alone cannot capture. 
However, this evidence in notes is frequently buried in lengthy, noisy text that is not intentionally written with any specific clinical prediction in mind. Summarization is an obvious mitigation, but generic summaries, tuned for fluency rather than the outcome, routinely omit decisive evidence while retaining plausible but uninformative detail. To this end, we propose \textbf{RASPER}, a \textbf{R}eward-\textbf{A}ligned \textbf{S}ummarizer for \textbf{P}rediction in \textbf{E}H\textbf{R}, that optimizes note summarization directly against the downstream clinical task. 
RASPER employs a tunable LLM-based summarizer to extract task-relevant evidence from discharge notes and trains it via reinforcement learning from prediction feedback, using a reward derived from the downstream predictor's loss.
To ground the summarizer, a longitudinal encoder converts structured codes into soft prompts that incorporate each patient's clinical context into note summarization.
By rewarding the quality of the resulting multimodal prediction, RASPER encourages the summarizer to retain patient-specific evidence that complements, rather than duplicates, information captured by structured codes.
RASPER consistently outperforms strong baselines on both readmission prediction and medication recommendation across MIMIC-III and MIMIC-IV.


\end{abstract}

\section{Introduction}
Electronic Health Records (EHRs) represent a patient's health trajectory with multimodal views: structured codes such as diagnoses capture standardized, cohort-level patterns across patients \cite{retain}, while discharge notes record patient-specific symptom severity, treatment response, and clinician observations that codes cannot fully record \cite{tayefi2021challenges}. 

Although discharge notes contain rich patient-specific information, they are primarily written for clinical documentation rather than predictive modeling. These raw notes are often lengthy, redundant, and sensitive to variation in documentation style \cite{liu2022note}. Directly feeding such notes into a model can therefore dilute the predictive signal and make the model sensitive to documentation artifacts rather than the underlying clinical state \cite{reprompt}. 
A natural strategy is to summarize discharge notes with an LLM ~\cite{healtllm1,healthllm2}, compressing noisy clinical narratives into a fluent, coherent summary before prediction.
However, reducing note length and improving readability does not by itself ensure that the retained information is useful for a specific prediction task.



The predictive value of a note summary depends on both the downstream task and the patient-specific context. Although different clinical prediction tasks may rely on overlapping clinical information, the information that is most predictive of each task can differ \cite{yoon2025aspect}. Hence, a summary optimized only for generic coverage and denoising is optimized for the wrong objective \cite{yadav2021reinforcement}, since it might discard details that a specific prediction depends on while preserving plausible but uninformative clinical information. This mismatch suggests that discharge note summarization should not be treated as a separate preprocessing step, but should instead be aligned with the downstream prediction and patient-specific objective.

Existing EHR-LLM methods do not directly address this problem. Some approaches feed raw or hard-prompted notes into LLMs and inherit the noise of the original documentation \cite{jiang2023health, huang2019clinicalbert}, and others rely on fixed summaries or extracted textual features that are never tuned to the target outcome \cite{reprompt}. Some structured code models discard the note modality altogether \cite{jiang2023graphcare, xu-etal-2024-ram}. As a result, prior methods generally do not align the summarization process with the downstream clinical objective.

To close this gap, we propose \textbf{RASPER}, a \textbf{R}eward-\textbf{A}ligned \textbf{S}ummarizer for \textbf{P}rediction in \textbf{E}H\textbf{R}. RASPER optimizes the summarization of notes directly against the downstream task: a tunable LLM summarizer is trained with reinforcement learning from prediction feedback (RLPF)~\cite{wu2025rlpf}, using the predictor's loss as the reward signal. A longitudinal EHR encoder maps structured codes into soft prompts that provide patient grounding. Because the reward reflects how well the note summary and the structured codes collaborate, the summarizer is encouraged to preserve note evidence that complements the structured code information and contributes to downstream prediction.
The summarizer is first initialized through supervised fine-tuning on teacher-generated summaries, then optimized with this prediction-feedback objective to preserve the task-relevant evidence that improves predictive utility.
Our approach makes the following contributions:

\begin{itemize}[leftmargin=*]





\item We propose a reward-aligned note summarization framework that trains a tunable LLM summarizer directly against the downstream prediction objective, using reinforcement learning from prediction feedback rather than treating summaries as a fixed preprocessing step.

\item We design an LLM-based predictor that jointly conditions on structured EHR soft prompts and generated note summaries, enabling prediction feedback to align the summarizer with both cohort-level code patterns and informative textual evidence.


\item We conduct experiments on MIMIC-III and MIMIC-IV, showing consistent improvements over strong deep-learning and LLM-based baselines on two clinical prediction tasks.

\end{itemize}
\section{Methodology}
\begin{figure*}[t]
    \centering
    \includegraphics[width=1\textwidth]{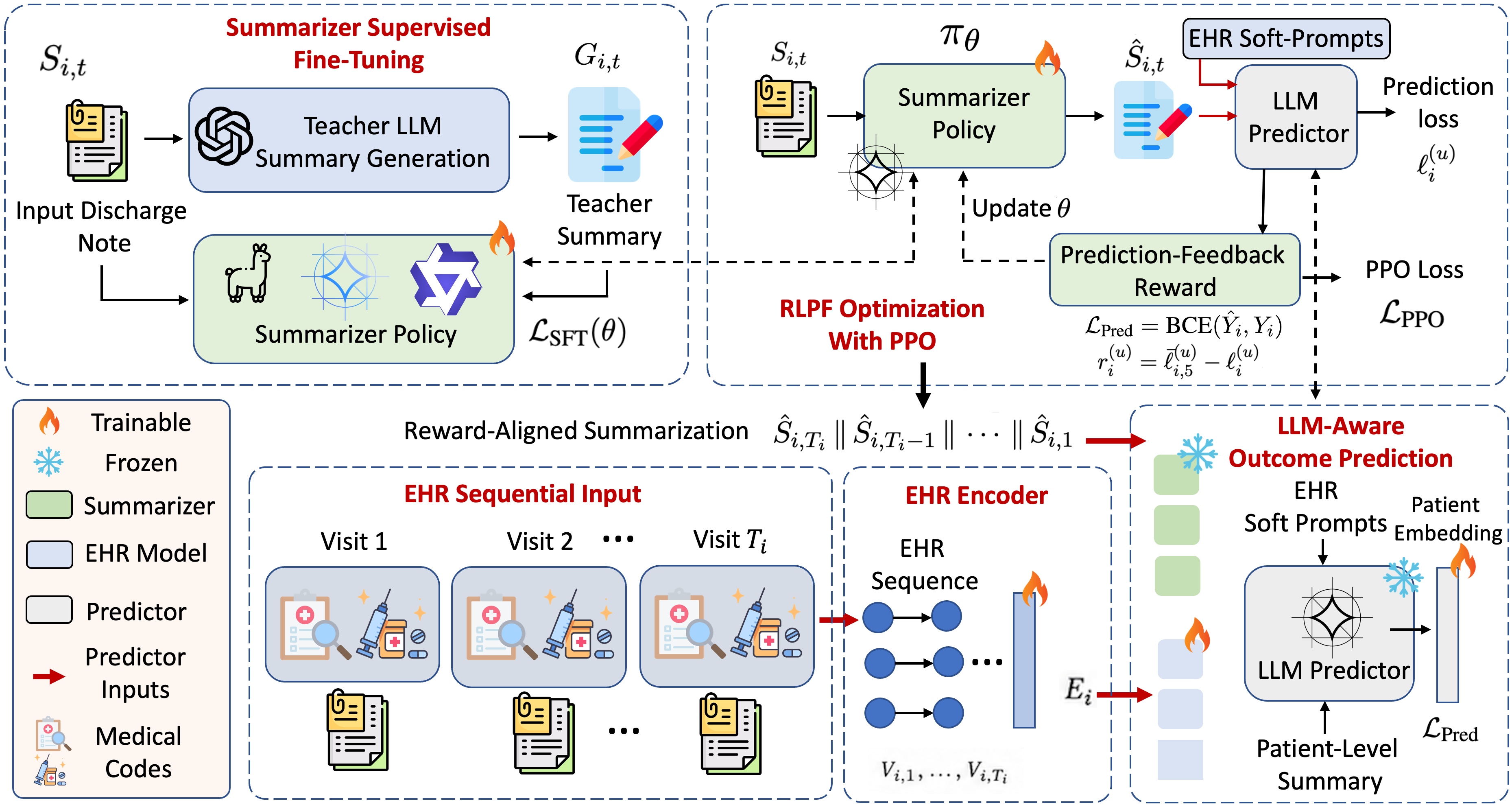}
    \caption{
        Overview of the proposed RASPER framework. A tunable LLM summarizer transforms discharge notes into standardized visit-level summaries, which are concatenated into a patient-level summary. An LLM-aware predictor combines this patient-level summary with soft prompt embeddings from structured EHR codes for outcome prediction. The summarizer is first initialized with supervised fine-tuning and then optimized with RLPF using prediction-loss-derived reward within the PPO objective.
    }
    \label{fig:methodology}
\end{figure*}
\subsection{Problem Formulation}
For each patient $i$, we use an ordered sequence of historical
clinical visits $\{V_{i,t}\}_{t=1}^{T_i}$ as input, where
$V_{i,t}$ denotes the $t$-th observed visit in chronological
order and $T_i$ denotes the number of historical visits
available before the target visit. Each visit $V_{i,t}$
contains a set of medical codes, including diagnoses,
medications, and procedures. Formally, the codes recorded at
visit $t$ are defined as $V_{i,t} = \{x^{i}_{j,t}\}_{j=1}^{|V_{i,t}|}$,
where $|V_{i,t}|$ denotes the number of medical codes associated with that visit. In addition to structured medical codes, each visit is accompanied by discharge notes represented
as text. We denote the token sequence of the discharge note for patient $i$ at visit $t$ as $S_{i,t} = \{s^{i}_{j,t}\}_{j=1}^{|S_{i,t}|}$, 
where each $s^{i}_{j,t}$ denotes a token and $|S_{i,t}|$ is the length of the note.

The target outcome is task-dependent and is defined as
\begin{equation}
Y_i \in
\begin{cases}
\{0,1\}, & \text{readmission prediction}\\
\{0,1\}^{K}, & \text{medication recommendation}
\end{cases}
\end{equation}
where $K$ denotes the size of the medication vocabulary. For medication recommendation, $Y_{i,k}=1$ indicates that
medication $k$ is prescribed at the target visit $T_i+1$.

\subsection{Model Overview}
Figure~\ref{fig:methodology} illustrates the overall architecture of \textbf{RASPER}. The model is designed around the idea that clinical-note summarization should not be optimized independently from the prediction task. The framework contains two interacting modules: an LLM-aware outcome predictor and a tunable LLM summarizer. The predictor combines structured EHR soft prompts with generated note summaries to model both cohort-level code patterns and patient-specific textual evidence. After supervised fine-tuning, the summarizer is further optimized through prediction feedback, so that its generated summaries preserve information that improves downstream clinical prediction.
\subsection{LLM-Aware Outcome Predictor}

Clinical prediction from EHR requires modeling both discharge notes and structured
medical codes \cite{reprompt}. Discharge notes describe patient condition,
treatment, and clinical progression \cite{MIMIC3}. In contrast,
diagnoses, procedures, and medications provide standardized cohort-level signals
\cite{retain, ARCI, gamenet1}. Using only notes may expose the model to
non-standardized textual evidence, and using only structured codes may miss
important patient-specific clinical context. Therefore, we integrate both in a unified representation space.

Large language models can encode clinical text into contextualized
representations \cite{gpt}, but many healthcare LLM approaches rely mainly on
hard prompting over linearized patient records \cite{healthllm3,healthllm7}. We
therefore use prompt tuning \cite{reprompt}, where the summarized discharge note
embeddings are augmented with soft prompts from a structured EHR encoder. This
allows the two modalities to complement each other in a lightweight and modular
prediction framework.

Mathematically, the input to the LLM for patient $i$ is defined as  
\begin{equation} 
\label{eq:llm_model} 
\boldsymbol{I}_{i}
=
[\boldsymbol{E}_{i}; \operatorname{Embed}(\hat{S}_{i})]
\in \mathbb{R}^{(P+C_i)\times D}.
\end{equation}  
The structured code sequence is first encoded by an encoder to obtain a patient-level representation \(\boldsymbol{h}^{\text{ehr}}_i\). For the EHR encoder, we use RETAIN \cite{retain}, a lightweight longitudinal model that captures visit-level and code-level patterns through attention (Further details on RETAIN are provided in Appendix \ref{app:retain}). We then project this representation into \(P\) continuous soft prompt embeddings:
\begin{equation}
\boldsymbol{E}_i = \operatorname{Proj}_{\text{ehr}}(\boldsymbol{h}^{\text{ehr}}_i),
\quad
\boldsymbol{E}_i \in \mathbb{R}^{P \times D}.
\end{equation}
where \(P\) is the number of soft prompt tokens and \(D\) is the hidden dimension of the LLM predictor. 

$\text{Embed}(\hat{S}_{i}) \in \mathcal{R}^{C_{i} \times D}$ denotes the token embeddings corresponding to the patient summary in the \textit{LLM-Aware Outcome Predictor}, where $C_{i}$ is the number of summary tokens for patient $i$. As defined in Equation \ref{eq:llm_embed_downstream}, the representation $I_{i}$ is then provided to the predictor LLM, which outputs the final hidden representation $H_{i} \in \mathbb{R}^{(P + C_{i}) \times D}$: 
\begin{equation} 
\label{eq:llm_embed_downstream} 
\boldsymbol{H}_{i} = \text{LLM}_{\text{pred}}(\boldsymbol{I}_{i}).
\end{equation}  

As defined in Equation \ref{eq:output}, we first apply average pooling over the final hidden
representations $H_i$ and then use a linear prediction head
to obtain task-specific logits:
\begin{equation}
\label{eq:output}
\mathbf{y}_{i} = \text{Linear}_{\text{pool}}(\boldsymbol{H}_{i}).
\end{equation}

The prediction loss is binary cross-entropy:
\begin{equation}
\mathcal{L}_{\mathrm{Pred}}
=
\mathrm{BCE}
\left(
\hat{\boldsymbol{Y}}_i,
Y_i
\right),
\end{equation}
where $\hat{Y}_i=\sigma(\mathbf{y}_i)$ denotes the predicted probability for patient $i$. The trainable components of the predictor include the EHR-to-soft-prompt layer, the prediction head, and the LoRA adapters of the LLM predictor. The backbone LLM parameters remain frozen.

\begin{figure}[t]
    \centering
    \includegraphics[width=0.47\textwidth]{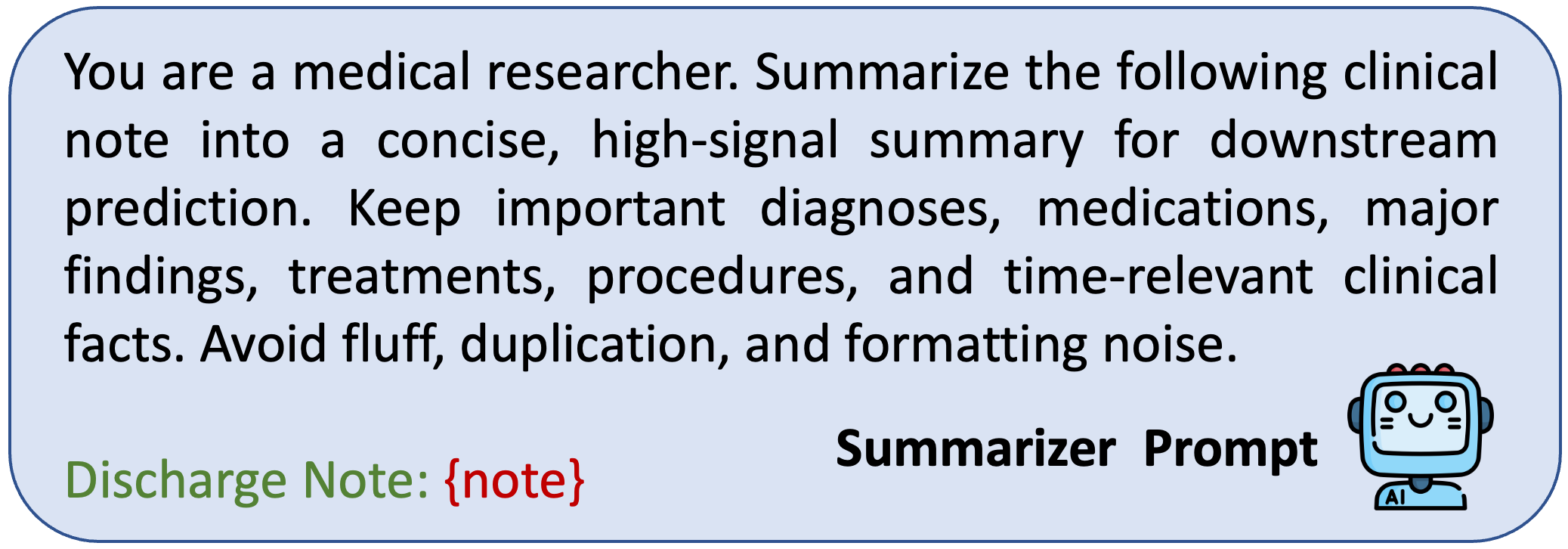}
    \caption{
        Prompt used by the tunable LLM summarizer to extract relevant information from discharge notes.
    }
    \label{fig:prompt}
\end{figure}
\subsection{Tunable LLM Summarizer}

The predictive value of clinical information depends on the downstream task and the patient-level prediction context. Hence, a generic summarization objective may not preserve the evidence most useful for outcome modeling \cite{wu2025rlpf}. This motivates using a summarizer that is not only a fixed preprocessing component, but can utilize prediction feedback from the downstream task.

The tunable LLM summarizer converts each raw discharge note into a shorter visit-level summary. It has two main goals: first, to reduce documentation noise and standardize the note format; second, to preserve patient-specific evidence that is useful for downstream prediction. After supervised initialization, the LLM-aware predictor provides a prediction-feedback reward by converting its per-sample prediction loss into a scalar reward, while the LLM summarizer is the policy model.

Given the discharge note $S_{i,t}$ for patient $i$ at time step $t$, the summarizer generates a standardized clinical summary $\hat{S}_{i,t}$ using the prompt shown in Figure~\ref{fig:prompt}. We then construct the patient-level document as $\hat{S}_i = \hat{S}_{i,T_i} \,\|\, \hat{S}_{i,T_i-1} \,\|\, \cdots \,\|\, \hat{S}_{i,1}$, where $\|$ denotes concatenation. This ordering places the most recent visit first because recent clinical status is often more directly related to near-future outcomes, while earlier visits provide additional historical context.
\vspace{-2pt}
\subsubsection{Summarizer Supervised Fine-Tuning}
Before applying reinforcement learning, we first perform supervised fine-tuning (SFT) to teach the local summarizer how to extract clinically relevant information from noisy discharge notes. 
Reinforcement-based alignment alone can be unstable when starting from an unadapted language model \cite{ouyang2022training}, especially in long-form clinical summarization settings where the output space is large, and rewards may be sparse or noisy \cite{rafailov2023direct}. Therefore, SFT first trains the model to produce summaries that follow a structured clinical style before the summarizer is optimized for downstream predictive utility.

Formally, for each visit-level discharge note $\boldsymbol{S}_{i,t}$, we first use the GPT-5.4 mini model to generate a target summary $G_{i,t}$, due to the lack of annotated visit-level summaries. This creates a supervised fine-tuning dataset
$\mathcal{D}_{\text{SFT}}=\{(S_{i,t}, G_{i,t})\}$, where $S_{i,t}$ is the input note and \(G_{i,t}\) is the teacher-generated target summary used for supervised fine-tuning.
We denote the summarizer policy as $\pi_{\theta}$, where $\theta$ represents the trainable LoRA adapter parameters \cite{hu2022lora}, while the backbone LLM weights remain frozen. The teacher-generated summary is written as
\begin{equation}
G_{i,t} = [g^1_{i,t}, g^2_{i,t}, \ldots, g^{L_{i,t}}_{i,t}],
\end{equation}
where $L_{i,t}$ denotes the length of the target summary for patient $i$ at visit $t$. At decoding step $\ell$, the summarizer predicts the next token $g^\ell_{i,t}$ conditioned on the discharge note $S_{i,t}$ and the previously generated target tokens $g^{<\ell}_{i,t}$. Therefore, the supervised fine-tuning objective is defined as the negative log-likelihood of the GPT-generated target summary:
\begin{equation}
\begin{aligned}
\mathcal{L}_{\text{SFT}}(\theta)
=
-\mathbb{E}_{(S_{i,t}, G_{i,t}) \sim \mathcal{D}_{\text{SFT}}}
\Bigg[
\sum_{\ell=1}^{L_{i,t}}
&\log \pi_{\theta}
\Big(
g^\ell_{i,t}
\mid \\
&S_{i,t}, g^{<\ell}_{i,t}
\Big)
\Bigg].
\end{aligned}
\end{equation}
By minimizing $\mathcal{L}_{\text{SFT}}(\theta)$ for several epochs, denoted by $\mathcal{Q}_{\text{SFT}}$, the summarizer learns to generate clinically meaningful summaries before being further optimized with the reinforcement learning objective.

\subsubsection{RLPF Optimization with PPO}

After supervised fine-tuning, we further optimize the summarizer using Reinforcement Learning from Predictor Feedback (RLPF) \cite{wu2025rlpf}. In this stage, the goal is to move beyond
generic clinical summarization and encourage the model to retain information that
improves downstream outcome prediction. The tunable summarizer is treated as a
policy model $\pi_{\theta}$ initialized from the SFT-trained checkpoint. Given the discharge note $\boldsymbol{S}_{i,t}$ for
patient $i$ at visit $t$, the policy generates a summary token-by-token:
\begin{equation}
\hat{S}_{i,t}
=
[\hat{s}^{1}_{i,t}, \hat{s}^{2}_{i,t}, \ldots, \hat{s}^{M_{i,t}}_{i,t}]
\sim
\pi_{\theta}(\cdot \mid S_{i,t}),
\end{equation}
where $M_{i,t}$ denotes the length of the generated summary. The generated
visit-level summaries are then concatenated to form the patient-level summary
$\hat{S}_{i}$, following the same construction described in the
previous subsection.

The LLM-aware predictor acts as the feedback model. Specifically, the
generated patient-level summary $\hat{S}_{i}$ is passed to the
LLM-aware outcome predictor defined in Equation~\ref{eq:llm_model}, together
with the structured EHR soft prompts. The predictor then produces the downstream
prediction score following Equations~\ref{eq:llm_embed_downstream}
and~\ref{eq:output}. We use the resulting prediction loss to define the reward
for optimizing the summarizer. This feedback helps the summarizer to produce
clinical summaries that are more useful for prediction. We optimize the summarizer using Proximal Policy Optimization (PPO)
\cite{schulman2017proximal}.

Let $\ell^{(u)}_i$ denote the per-sample binary cross-entropy loss for patient
$i$ at epoch $u$:
\begin{equation}
\ell^{(u)}_i
=
\text{BCE}(\boldsymbol{Y}^{(u)}_i, Y_i).
\end{equation}
For each patient, we maintain a history of previous losses during warm-up in each epoch and define a strong
reference loss as the average of the five lowest losses observed so far:
\begin{equation}
\bar{\ell}^{(u)}_{i,5}
=
\frac{1}{5}
\sum_{m=1}^{5}
\ell^{(u,m)}_{i,\text{lowest}},
\end{equation}
where $\ell^{(u,m)}_{i,\text{lowest}}$ denotes the $m$-th smallest historical
loss for patient $i$ observed during the LLM predictor warm-up stage, before the summarizer is optimized. Using the average of the lowest historical losses provides a reference of the strongest prediction performance observed for that patient during warm-up, and it makes the reward less sensitive to differences across patients. The raw reward is computed as the improvement over this
reference:
\begin{equation}
r^{(u)}_i
=
\bar{\ell}^{(u)}_{i,5}
-
\ell^{(u)}_i.
\end{equation}

This reward is used as the prediction-feedback signal for PPO optimization of the summarizer policy; a higher reward indicates that the generated summary improves downstream prediction relative to the patient's warm-up best historical baselines.

To stabilize PPO training, the patient-level rewards are normalized within each
mini-batch $\mathcal{B}$. Let $\mu_{\mathcal{B}}$ and $\sigma_{\mathcal{B}}$
denote the mean and standard deviation of rewards in the mini-batch. The
normalized reward is
\begin{equation}
\tilde{r}^{(u)}_i
=
\frac{
r^{(u)}_i - \mu_{\mathcal{B}}
}{
\sigma_{\mathcal{B}} + \epsilon
},
\end{equation}
where $\epsilon$ is a small constant for numerical stability. The normalized
reward is then clipped to avoid unstable updates:
\begin{equation}
\hat{r}^{(u)}_i
=
\text{clip}
\left(
\tilde{r}^{(u)}_i,
-r_{\max},
r_{\max}
\right).
\end{equation}

Because the reward is computed at the patient level, the same reward is assigned
to all generated visit summaries belonging to that patient. For each generated
summary token, we compute the log-probability under the current policy and under
the old policy that produced the sampled summary. Let
$\pi_{\theta_{\text{old}}}$ denote the policy before the PPO update. The PPO
probability ratio for token $\hat{s}^{\ell}_{i,t}$ is
\begin{equation}
\rho^{\ell}_{i,t}(\theta)
=
\frac{
\pi_{\theta}
\left(
\hat{s}^{\ell}_{i,t}
\mid
S_{i,t}, \hat{s}^{<\ell}_{i,t}
\right)
}{
\pi_{\theta_{\text{old}}}
\left(
\hat{s}^{\ell}_{i,t}
\mid
S_{i,t}, \hat{s}^{<\ell}_{i,t}
\right)
}.
\end{equation}

In addition to the policy model, we use a value head $V_{\phi}$ on top of the
summarizer hidden states to estimate the expected reward for each generated
token. The token-level advantage is computed as
\begin{equation}
A^{\ell}_{i,t}
=
\hat{r}^{(u)}_i
-
V_{\phi}
\left(
\boldsymbol{h}^{\ell}_{i,t}
\right),
\end{equation}
where $\boldsymbol{h}^{\ell}_{i,t}$ is the hidden state corresponding to token
$\hat{s}^{\ell}_{i,t}$. The PPO clipped policy loss is then defined as
\begin{equation}
\begin{aligned}
\mathcal{L}_{\text{policy}}
= -\mathbb{E}\Big[
\min\Big(
\rho^{\ell}_{i,t}(\theta) A^{\ell}_{i,t},
\operatorname{clip}\Big(
\rho^{\ell}_{i,t}(\theta), \\
\qquad 1-\epsilon_{\text{PPO}},
1+\epsilon_{\text{PPO}}
\Big)
A^{\ell}_{i,t}
\Big)
\Big].
\end{aligned}
\end{equation}
where $\epsilon_{\text{PPO}}$ is the PPO clipping threshold.

The value head is trained to regress to the normalized and clipped reward:
\begin{equation}
\mathcal{L}_{\text{value}}
=
\mathbb{E}
\left[
\left(
V_{\phi}
\left(
\boldsymbol{h}^{\ell}_{i,t}
\right)
-
\hat{r}^{(u)}_i
\right)^2
\right].
\end{equation}

To prevent the policy from drifting too far from the SFT-initialized behavior,
we use a frozen reference policy $\pi_{\text{ref}}$. We use a sampled KL penalty over the generated tokens to discourage the current policy from drifting too far from the frozen SFT reference policy \cite{ouyang2022training}:
\begin{equation}
\begin{aligned}
\mathcal{L}_{\text{KL}}
=
\mathbb{E}
\Big[
&\log
\pi_{\theta}
\big(
\hat{s}^{\ell}_{i,t}
\mid
S_{i,t}, \hat{s}^{<\ell}_{i,t}
\big) \\
&-
\log
\pi_{\text{ref}}
\big(
\hat{s}^{\ell}_{i,t}
\mid
S_{i,t}, \hat{s}^{<\ell}_{i,t}
\big)
\Big].
\end{aligned}
\end{equation}
We also include an entropy bonus to encourage sufficient exploration during
summary generation \cite{zheng2023secrets}:
\begin{equation}
\mathcal{H}_{i,t}^{\ell}
=
-
\sum_{v \in \mathcal{V}}
\pi_{\theta}
\left(
v
\mid
S_{i,t}, \hat{s}^{<\ell}_{i,t}
\right)
\log
\pi_{\theta}
\left(
v
\mid
S_{i,t}, \hat{s}^{<\ell}_{i,t}
\right),
\end{equation}
where $\mathcal{V}$ is the vocabulary.

The final PPO objective combines the clipped policy loss, value loss, entropy
regularization, and KL regularization:
\begin{equation}
\mathcal{L}_{\text{PPO}}
=
\mathcal{L}_{\text{policy}}
+
\lambda_{v}\mathcal{L}_{\text{value}}
-
\lambda_{H}\mathbb{E}[\mathcal{H}_{i,t}^{\ell}]
+
\lambda_{\text{KL}}\mathcal{L}_{\text{KL}},
\end{equation}
where $\lambda_v$, $\lambda_H$, and $\lambda_{\text{KL}}$ control the
contributions of the value loss, entropy bonus, and KL penalty, respectively.

\subsection{Staged Predictor–Summarizer Optimization}
The RL optimization stage requires a reliable downstream predictor to provide meaningful
feedback for the summarizer. When the predictor is under-trained, the reward
signal can be noisy and may lead to unstable policy updates. To address this, we
first train the predictor for $Q_{\text{warm}}$ epochs using
$\mathcal{L}_{\text{Pred}}$, and then optimize the summarizer using
$\mathcal{L}_{\text{PPO}}$, while the backbone of the predictor LLM is frozen. The two modules use separate optimizers, $\mathcal{O}_{\text{Pred}}$ and $\mathcal{O}_{\text{PPO}}$:
\begin{equation}
\begin{cases}

\mathcal{O}_{\text{Pred}}(\mathcal{L}_{\text{Pred}}),
& e < Q_{\text{warm}} \\[9pt]
\begin{aligned}
\mathcal{O}_{\text{Pred}}(\mathcal{L}_{\text{Pred}}),\\

\mathcal{O}_{\text{PPO}}(\mathcal{L}_{\text{PPO}})
\end{aligned}
& e \ge Q_{\text{warm}}
\end{cases}
\end{equation}
After the warm-up stage, the predictor and summarizer are
optimized with separate objectives and optimizers. The
predictor continues minimizing
$\mathcal{L}_{\mathrm{Pred}}$, while the summarizer is updated
using $\mathcal{L}_{\mathrm{PPO}}$ based on the
prediction-feedback reward. This staged training provides a more reliable reward signal before the summarizer policy is updated, and reduces the risk of optimizing the summarizer against under-trained predictor feedback.

\section{Evaluation}
\subsection{Datasets and Setups}
In this study, we use two real-world datasets to evaluate model performance on medication and readmission prediction tasks.

\begin{table}[t]
\centering
\caption{Dataset statistics.}
\label{Statistical}
\small
\renewcommand{\arraystretch}{1.1}
\setlength{\tabcolsep}{3pt} 
\makebox[\linewidth]{
    \begin{tabular}{@{}lcc@{}}
    \hline
    Metric & \multicolumn{1}{c}{\bf{MIMIC-III}} & \multicolumn{1}{c}{\bf{MIMIC-IV}}\\
    \hline
    \# of patients & 7537 & 15874  \\
    \# of medications per patient  & 79.30 & 113.55  \\
    \# of diagnoses per patient & 28.98  & 65.99  \\
    \# of procedures per patient & 7.37 & 9.08  \\
    \# of history visits per patient   & 1.65 & 3.11  \\
    Positive rate for readmission &  53.7\% &  53.5\% \\
    \hline
    \end{tabular}
}
\vspace{-0.2cm}
\end{table}

\noindent $\bullet$ \textbf{MIMIC-III} \cite{MIMIC3} is an open-access database containing health records for over 40,000 patients admitted to critical care units between 2001 and 2012. In this study, we focus on patients with multiple visits and aim to predict binary outcomes.\\
\noindent $\bullet$ \textbf{MIMIC-IV} \cite{johnson2020mimic} is a publicly available EHR dataset containing records from over 380,000 patients admitted to Beth Israel Deaconess Medical Center between 2008 and 2019. It extends MIMIC-III with a clearer modular structure, richer clinical detail, and improved data provenance.

Both MIMIC-III and MIMIC-IV are publicly available through PhysioNet under controlled-access terms for credentialed researchers. The dataset statistics are summarized in Table \ref{Statistical}. In this study, we consider two key healthcare prediction tasks: hospital readmission prediction (binary classification) and medication recommendation (multi-label classification). Given patient information up to visit $T_i$, the model predicts the outcome at visit $T_{i+1}$ for patient $i$ based on $T_i$ prior visits.
Since these tasks require longitudinal data, we excluded patients with only a single recorded visit. We randomly split the data into 70\% training and 30\% testing sets and report the mean over 4 seeds. The source code is released.\footnote{Code: \url{https://github.com/aryahm1375/RASPER}.} 

\begin{table}[t]
\centering
\small
\caption{Hyperparameters used in the proposed framework.}
\label{tab:hyperparameters}
\setlength{\tabcolsep}{3.5pt}
\begin{tabular}{lcc}
\toprule
Component & Hyperparameter / Symbol & Value \\
\midrule
Soft prompt encoder & $P$ & 8 \\
LLM hidden size & $D$ & 2304 \\
Max Summary length & $C_i$ & 512 \\
Reward normalization & $\epsilon$ & $1\times10^{-8}$ \\
Reward clipping & $r_{\max}$ & 2.0 \\
PPO clipping & $\epsilon_{\text{PPO}}$ & 0.2 \\
Value loss weight & $\lambda_v$ & 0.5 \\
Entropy weight & $\lambda_H$ & 0.01 \\
KL weight & $\lambda_{\text{KL}}$ & 0.02 \\
Max PPO generation & $M_{i,t}$ & 128 \\
SFT epochs & $Q_{\mathrm{SFT}}$ & 10 \\
Predictor warm-up & $Q_{\text{warm}}$ & 20 \\
Optimizers & $\mathcal{O}_{\text{Pred}}, \mathcal{O}_{\text{PPO}}$ & AdamW \\
\bottomrule
\end{tabular}
\end{table}

\begin{table*}[t]
\centering
\small
\caption{Baseline performance comparison on MIMIC-III and MIMIC-IV. Best results are in bold, and underlined results denote the second-best results that are statistically lower at the 95\% confidence level.}
\label{tab:main_results}
\setlength{\tabcolsep}{4.5pt}
\renewcommand{\arraystretch}{1.12}
\begin{tabular}{l|cc|cc||cc|cc}
\toprule
\rowcolor{gray!15}
&
\multicolumn{4}{c||}{\textbf{MIMIC-III}}
& \multicolumn{4}{c}{\textbf{MIMIC-IV}} \\

\rowcolor{gray!15}
\multicolumn{1}{c|}{\cellcolor{gray!15}\textbf{Model}}
&
\multicolumn{2}{c|}{\textbf{Readmission}}
& \multicolumn{2}{c||}{\textbf{Medication}}
& \multicolumn{2}{c|}{\textbf{Readmission}}
& \multicolumn{2}{c}{\textbf{Medication}} \\

\rowcolor{gray!15}
&
\textbf{AUROC} & \textbf{PRAUC}
& \textbf{F1} & \textbf{Jaccard}
& \textbf{AUROC} & \textbf{PRAUC}
& \textbf{F1} & \textbf{Jaccard} \\
\midrule
\rowcolor{green!12}
\multicolumn{9}{c}{\textbf{DL-based Methods}} \\
\midrule
Deepr \cite{nguyen2016mathtt}     & 0.671 & 0.702 & 0.255 & 0.160 & 0.621 & 0.651 & 0.228 & 0.189 \\
RETAIN \cite{retain}    & 0.657 & 0.678 & 0.253 & 0.156 & 0.673 & 0.698 & 0.349 & 0.255 \\
GRASP \cite{zhang2021grasp}     & 0.618 & 0.635 & 0.153 & 0.096 & 0.535 & 0.578 & 0.267 & 0.195 \\
AdaCare \cite{ma2020adacare}   & 0.645 & 0.687 & 0.235 & 0.133 & 0.615 & 0.644 & 0.251 & 0.184 \\
StageNet \cite{gao2020stagenet}  & 0.678 & 0.705 & 0.239 & 0.139 & 0.659 & 0.690 & 0.278 & 0.202 \\
Adore \cite{cheong2023adaptive}     & 0.665 & 0.698 & 0.152 & 0.125 & 0.615 & 0.643 & 0.206 & 0.165 \\
ARCI \cite{ARCI}      & 0.648 & 0.669 & 0.283 & 0.184 & 0.668 & 0.698 & 0.303 & 0.246 \\

\midrule
\rowcolor{blue!12}
\multicolumn{9}{c}{\textbf{LLM-based Methods}} \\
\midrule
GraphCare \cite{jiang2023graphcare} & 0.665 & 0.705 & 0.294 & 0.195 & 0.688 & 0.711 & 0.355 & 0.259 \\
LINKO \cite{kerdabadi2025multi}     & 0.674 & 0.711 & 0.304 & 0.202 & 0.695 & 0.715 & 0.367 & 0.264 \\
RePrompT \cite{reprompt}  & \underline{0.682} & \underline{0.718} & \underline{0.312} & \underline{0.206}
          & \underline{0.702} & \underline{0.723} & \underline{0.375} & \underline{0.273} \\

\midrule
\rowcolor{gray!7}
\textbf{RASPER}
          & \textbf{0.704} & \textbf{0.738} & \textbf{0.326} & \textbf{0.218}
          & \textbf{0.725} & \textbf{0.742} & \textbf{0.388} & \textbf{0.282} \\

\bottomrule
\end{tabular}
\end{table*}

\begin{table*}[t]
\centering
\small
\caption{Ablation studies on MIMIC-III and MIMIC-IV. Best results are in bold, and underlined results denote the second-best results that are statistically lower at the 95\% confidence level.}
\label{tab:ablation}
\setlength{\tabcolsep}{4.5pt}
\renewcommand{\arraystretch}{1.12}
\begin{tabular}{l|cc|cc||cc|cc}
\toprule
\rowcolor{gray!15}
&
\multicolumn{4}{c||}{\textbf{MIMIC-III}}
& \multicolumn{4}{c}{\textbf{MIMIC-IV}} \\
\rowcolor{gray!15}
\multicolumn{1}{c|}{\cellcolor{gray!15}\textbf{Model Variant}}
&
\multicolumn{2}{c|}{\textbf{Readmission}}
& \multicolumn{2}{c||}{\textbf{Medication}}
& \multicolumn{2}{c|}{\textbf{Readmission}}
& \multicolumn{2}{c}{\textbf{Medication}} \\
\rowcolor{gray!15}
&
\textbf{AUROC} & \textbf{PRAUC}
& \textbf{F1} & \textbf{Jaccard}
& \textbf{AUROC} & \textbf{PRAUC}
& \textbf{F1} & \textbf{Jaccard} \\
\midrule
RASPER w/o EHR encoder
& 0.661 & 0.698 & 0.288 & 0.187 
& 0.685 & 0.704 & 0.353 & 0.258  \\
RASPER w/o Summarization
& 0.674 & 0.712 & 0.298 & 0.191
& 0.691 & 0.713 & 0.363 & 0.262 \\

RASPER w/o RL (GPT only)
& \underline{0.689} & \underline{0.725} & \underline{0.309} & \underline{0.201}
& \underline{0.708} & \underline{0.723} & \underline{0.371} & \underline{0.272} \\

RASPER w/o RL (SFT only)
& 0.684 & 0.721 & 0.305 & 0.198
& 0.701 & 0.718 & 0.367 & 0.269 \\

\midrule
\rowcolor{gray!7}
\textbf{RASPER}
& \textbf{0.704} & \textbf{0.738} & \textbf{0.326} & \textbf{0.218}
& \textbf{0.725} & \textbf{0.742} & \textbf{0.388} & \textbf{0.282} \\

\bottomrule
\end{tabular}
\end{table*}
For both the LLM summarizer and the LLM-aware predictor, we utilize the Gemma-2-2B model. All experiments are implemented in Python, using PyTorch \cite{pytorch} as the main deep learning framework. RASPER is also fully compatible with the PyHealth framework \cite{pyhealth}, which we use for implementing EHR baselines. For LLM tuning, we use the Hugging Face framework \cite{wolf2019huggingface}. Experiments are conducted on a high-performance server with three NVIDIA A6000 GPUs. 

The hyperparameters used in our experiments are shown in Table~\ref{tab:hyperparameters}. These values were selected through a greedy search for the best performance.



\subsection{Baselines}
\vspace{-0.2cm}

We compare RASPER with representative deep EHR models, including Deepr~\cite{nguyen2016mathtt}, RETAIN~\cite{retain}, GRASP~\cite{zhang2021grasp}, AdaCare~\cite{ma2020adacare}, StageNet~\cite{gao2020stagenet}, ADORE~\cite{cheong2023adaptive}, and ARCI~\cite{ARCI}. We also compare against recent LLM-based healthcare prediction methods, including GraphCare~\cite{jiang2023graphcare}, LINKO~\cite{kerdabadi2025multi}, and RePrompT~\cite{reprompt}. These baselines cover sequential EHR modeling, ontology-enhanced prediction, patient-similarity learning, knowledge-graph augmentation, and LLM-based EHR representation learning. More information about baselines is available in Appendix \ref{app:baselines}.

\begin{figure*}[t]
    \centering

    \begin{subfigure}{0.95\textwidth}
        \centering
        \includegraphics[width=\linewidth]{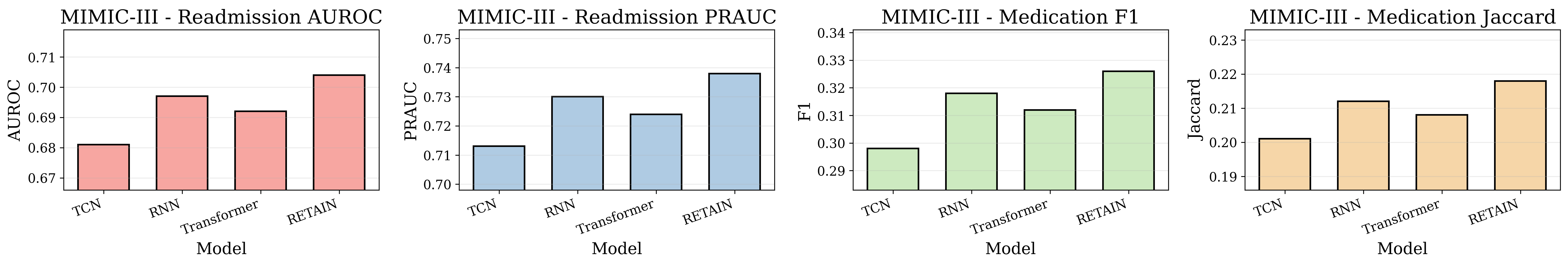}
        
        \label{fig:mimic_iii}
    \end{subfigure}
    \hfill
    \begin{subfigure}{0.95\textwidth}
        \centering
        \includegraphics[width=\linewidth]{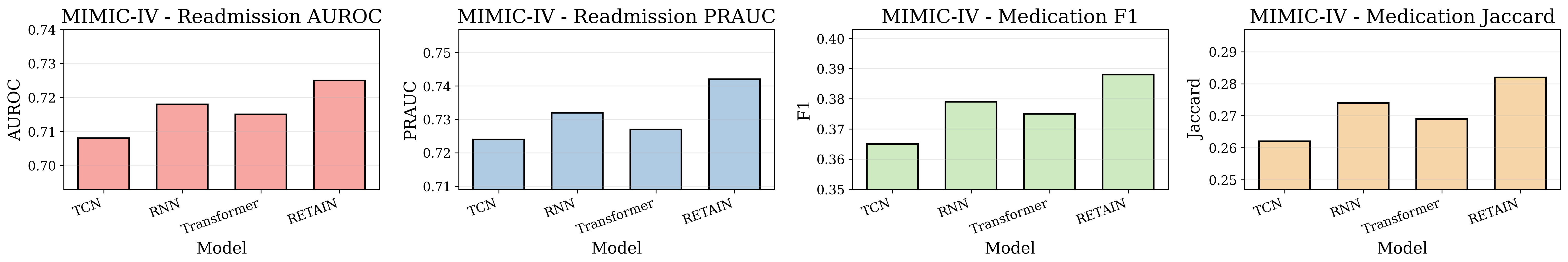}
        
        \label{fig:mimic_iv}
    \end{subfigure}

    \caption{Performance comparison across different EHR models on MIMIC-III and MIMIC-IV.}
    \label{fig:mimic_comparison}
\end{figure*}

\begin{table*}[t]
\centering
\small
\caption{Comparison of LLM backbones on MIMIC-III and MIMIC-IV across two tasks.}
\label{tab:llm_selection}
\setlength{\tabcolsep}{4.5pt}
\renewcommand{\arraystretch}{1.12}
\begin{tabular}{l|cc|cc||cc|cc}
\toprule
\rowcolor{gray!15}
&
\multicolumn{4}{c||}{\textbf{MIMIC-III}}
& \multicolumn{4}{c}{\textbf{MIMIC-IV}} \\
\rowcolor{gray!15}
\multicolumn{1}{c|}{\cellcolor{gray!15}\textbf{Backbone}}
&
\multicolumn{2}{c|}{\textbf{Readmission}}
& \multicolumn{2}{c||}{\textbf{Medication}}
& \multicolumn{2}{c|}{\textbf{Readmission}}
& \multicolumn{2}{c}{\textbf{Medication}} \\
\rowcolor{gray!15}
&
\textbf{AUROC} & \textbf{PRAUC}
& \textbf{F1} & \textbf{Jaccard}
& \textbf{AUROC} & \textbf{PRAUC}
& \textbf{F1} & \textbf{Jaccard} \\
\midrule

Llama-3.2-1B
& 0.701 & 0.734 & 0.321 & 0.214
& 0.722 & 0.737 & 0.384 & 0.278 \\

Qwen2.5-0.5B
& 0.697 & 0.730 & 0.315 & 0.209
& 0.716 & 0.731 & 0.379 & 0.276 \\

\midrule
\rowcolor{gray!7}
\textbf{Gemma-2-2B}
& \textbf{0.704} & \textbf{0.738} & \textbf{0.326} & \textbf{0.218}
& \textbf{0.725} & \textbf{0.742} & \textbf{0.388} & \textbf{0.282} \\

\bottomrule
\end{tabular}
\end{table*}

\subsection{Discussion}
\subsubsection{Performance Comparison on Baselines}
As shown in Table \ref{tab:main_results}, the proposed method achieves the best performance across the baselines. RePrompT is the closest competing method, which is expected since it also uses a multimodal LLM-based design. The improvement over RePrompT demonstrates the effectiveness
of the overall RASPER framework for jointly leveraging
structured EHR representations and prediction-aligned
clinical summaries. The gap between RETAIN, RePrompT, and our method also shows the value of using clinical notes in EHR modeling. Compared with models that rely only on structured codes, adding note information clearly improves the learned patient representations. In general, the LLM-based methods perform better than traditional deep learning baselines. Even when prior methods do not directly use clinical-note summarization, they suggest that LLM-based feature extraction~\cite{jiang2023graphcare, kerdabadi2025multi} and multimodal learning~\cite{reprompt} are useful for prediction.
\subsubsection{Ablation Studies on Modules}
As shown in Table~\ref{tab:ablation}, removing the EHR encoder leads to the largest performance drop, indicating that cohort-level structured information is essential for complementing patient-specific clinical notes. In addition, directly using noisy clinical notes as input leads to worse performance than the proposed method, supporting the main motivation of our paper: raw discharge notes contain redundant and noisy information that can distract the predictor. The GPT-based summarizer improves over raw-note input by reducing this noise, but it still relies on generic summarization objectives and is not explicitly optimized for the clinical prediction task. In contrast, our proposed method further improves performance by aligning the summaries with downstream prediction loss through RL-based optimization. This gap shows that prediction-aligned summarization adds value beyond generic note compression and helps preserve useful information for the final prediction.

\subsubsection{Ablation Studies on Encoder Selection}
The EHR encoder plays a central role in converting the structured medical codes in a patient’s longitudinal history into cohort-level representations. To study this effect, we replace the structured encoder with different EHR modeling backbones and report the results in Figure \ref{fig:mimic_comparison}. Among the compared encoders, RETAIN achieves the best overall performance. This is likely because RETAIN captures patient history at multiple levels. As a result, the generated soft tokens provide more informative structured signals to the LLM predictor. The gains are consistent across datasets and evaluation metrics, suggesting that the benefit is not limited to a single task setting. These results also indicate that the proposed framework can benefit from stronger structured EHR encoders.

\subsubsection{LLM Model Selection}

To examine the sensitivity of RASPER to the choice of LLM backbone, we conduct additional experiments using different LLMs, as shown in Table~\ref{tab:llm_selection}. Since RASPER relies on LLMs for both clinical note summarization and prediction, changing the backbone can affect the final performance. Overall, Gemma-2-2B achieves the best performance across all metrics, suggesting that a stronger backbone can provide better clinical representations and improve downstream prediction. However, the performance gaps between Gemma-2-2B, Llama-3.2-1B, and Qwen2.5-0.5B are relatively small in most settings. This indicates that RASPER is not highly dependent on a specific LLM architecture, and the main performance gain comes from the proposed framework.
\section{Related Work}
A large body of work has studied how to learn predictive representations from Electronic Health Records for clinical decision support. Early neural models such as RETAIN \cite{retain} improve interpretability by applying reverse-time attention over visits and medical variables, allowing the model to emphasize clinically important parts of a patient's history. Other methods incorporate medical ontologies to reduce sparsity and improve the use of domain knowledge. For example, Hierarchical Attention Propagation (HAP) \cite{hap} propagates attention through ontology structures to capture relationships between ancestor and descendant concepts. Temporal disease progression and patient heterogeneity have also been widely studied. AdaCare \cite{ma2020adacare} models both short- and long-term variations in clinical signals using multi-scale convolutional modules, while StageNet \cite{gao2020stagenet} introduces stage-aware modeling to capture disease progression over time. These methods improve clinical prediction by better modeling structured EHR sequences, attention patterns, or disease stages. However, they mainly rely on structured medical codes or numerical features, and therefore do not fully exploit the rich information in unstructured clinical notes.

Recent advances in Large Language Models (LLMs) have motivated their use for EHR representation learning and clinical prediction. GraphCare \cite{jiang2023graphcare} constructs patient-specific knowledge graphs by combining structured knowledge bases with LLM-generated knowledge, and applies a bi-attention-augmented GNN for prediction. RAM-EHR \cite{xu-etal-2024-ram} augments patient representations through LLM-powered dense retrieval from multiple knowledge sources, together with consistency regularization to improve robustness. Prompting-based studies \cite{zhu2024prompting} further show that carefully designed prompts incorporating EHR-specific information can enable LLMs to perform clinical prediction in few-shot or zero-shot settings. LINKO \cite{kerdabadi2025multi} uses LLM-initialized medical concept embeddings and performs dual-axis knowledge propagation, both within individual ontologies and across different ontology systems, and RePrompT \cite{reprompt} incorporates structured EHR encoders into LLMs through prompt tuning, using recurrent visit-level states and cohort-trained soft prompts. 
\section{Conclusion}
We proposed \textbf{RASPER}, a reward-aligned framework that integrates structured EHR codes and discharge notes for clinical prediction. RASPER uses EHR soft prompts to capture cohort-level patterns and prediction-aligned summaries to retain useful patient-specific context. By optimizing the summarizer with downstream prediction feedback, RASPER produces summaries that are more directly useful for the target task. Experiments on MIMIC-III and MIMIC-IV show consistent improvements over strong baselines.

\section{Limitations}
Although RASPER shows strong performance, this study has two limitations. First, the experiments are conducted only on MIMIC-III and MIMIC-IV, so additional evaluation on datasets from other hospitals would better show its generalizability. Second, the current framework focuses on prediction performance and does not deeply analyze how different summary lengths or prompt formats affect the final results. 
\section{Ethical Considerations}
This study uses publicly available, de-identified EHR datasets and does not involve direct interaction with patients. The proposed framework is intended to support clinical prediction research rather than replace clinical judgment. Because generated summaries may omit clinically relevant
information or introduce inaccuracies, the proposed
framework should not be used as an autonomous clinical
decision-making system. We used AI-assisted tools to support language editing, grammar correction, and clarity improvement during manuscript preparation.
\section{Acknowledgments}
This work was supported by the National Science Foundation under Grant No. IIS-2544634.


\bibliography{references}

@inproceedings{gamenet1,
  title={Gamenet: Graph augmented memory networks for recommending medication combination},
  author={Shang, Junyuan and Xiao, Cao and Ma, Tengfei and Li, Hongyan and Sun, Jimeng},
  booktitle={proceedings of the AAAI Conference on Artificial Intelligence},
  volume={33},
  pages={1126--1133},
  year={2019}
}

@article{retain,
  title={Retain: An interpretable predictive model for healthcare using reverse time attention mechanism},
  author={Choi, Edward and Bahadori, Mohammad Taha and Sun, Jimeng and Kulas, Joshua and Schuetz, Andy and Stewart, Walter},
  journal={Advances in neural information processing systems},
  volume={29},
  year={2016}
}

@article{MIMIC3,
  title={MIMIC-III, a freely accessible critical care database},
  author={Johnson, Alistair EW and Pollard, Tom J and Shen, Lu and Lehman, Li-wei H and Feng, Mengling and Ghassemi, Mohammad and Moody, Benjamin and Szolovits, Peter and Anthony Celi, Leo and Mark, Roger G},
  journal={Scientific data},
  volume={3},
  number={1},
  pages={160035},
  year={2016},
  publisher={Nature Publishing Group}
}

@article{pytorch,
  title={Pytorch: An imperative style, high-performance deep learning library},
  author={Paszke, Adam and Gross, Sam and Massa, Francisco and Lerer, Adam and Bradbury, James and Chanan, Gregory and Killeen, Trevor and Lin, Zeming and Gimelshein, Natalia and Antiga, Luca and Desmaison, Alban and K{\"o}pf, Andreas and Yang, Edward and DeVito, Zach and Raison, Martin and Tejani, Alykhan and Chilamkurthy, Sasank and Steiner, Benoit and Fang, Lu and Bai, Junjie and Chintala, Soumith},
  journal={Advances in neural information processing systems},
  volume={32},
  year={2019}
}

@inproceedings{pyhealth,
    author = {Yang, Chaoqi and Wu, Zhenbang and Jiang, Patrick and Lin, Zhen and Gao, Junyi and Danek, Benjamin P. and Sun, Jimeng},
title = {PyHealth: A Deep Learning Toolkit for Healthcare Applications},
year = {2023},
isbn = {9798400701030},
publisher = {Association for Computing Machinery},
address = {New York, NY, USA},
booktitle = {Proceedings of the 29th ACM SIGKDD Conference on Knowledge Discovery and Data Mining},
pages = {5788–5789},
numpages = {2},
location = {Long Beach, CA, USA},
series = {KDD '23}
}

@inproceedings{jiang2023graphcare,
  title={Graphcare: Enhancing healthcare predictions with personalized knowledge graphs},
  author={Jiang, Pengcheng and Xiao, Cao Danica and Cross, Adam and Sun, Jimeng},
  booktitle={International Conference on Learning Representations},
  volume={2024},
  pages={16839--16866},
  year={2024}
}

@inproceedings{ARCI,
  title={Contrastive learning on medical intents for sequential prescription recommendation},
  author={H Moghaddam, Arya and Nayebi Kerdabadi, Mohsen and Liu, Mei and Yao, Zijun},
  booktitle={Proceedings of the 33rd ACM International Conference on Information and Knowledge Management},
  pages={748--757},
  year={2024}
}

@article{johnson2020mimic,
    author = {Johnson, Alistair and Bulgarelli, Lucas and Pollard, Tom and Horng, Steven and Celi, Leo Anthony and Mark, Roger},
  title = {{MIMIC-IV}},
  journal = {{PhysioNet}},
  year = {2021},
  month = mar,
  
}

@article{nguyen2016mathtt,
  title={Deepr: a convolutional net for medical records},
  author={Nguyen, Phuoc and Tran, Truyen and Wickramasinghe, Nilmini and Venkatesh, Svetha},
  journal={IEEE journal of biomedical and health informatics},
  volume={21},
  number={1},
  pages={22--30},
  year={2017},
  publisher={IEEE}
}

@inproceedings{healtllm1,
  title={Healai: A healthcare llm for effective medical documentation},
  author={Goyal, Sagar and Rastogi, Eti and Rajagopal, Sree Prasanna and Yuan, Dong and Zhao, Fen and Chintagunta, Jai and Naik, Gautam and Ward, Jeff},
  booktitle={Proceedings of the 17th ACM International Conference on Web Search and Data Mining},
  pages={1167--1168},
  year={2024}
}

@inproceedings{healthllm2,
  title={Llm-based framework for administrative task automation in healthcare},
  author={Gebreab, Senay A and Salah, Khaled and Jayaraman, Raja and ur Rehman, Muhammad Habib and Ellaham, Samer},
  booktitle={2024 12th International Symposium on Digital Forensics and Security (ISDFS)},
  pages={1--7},
  year={2024},
  organization={IEEE}
}

@article{healthllm3,
  title={Prompt engineering as an important emerging skill for medical professionals: tutorial},
  author={Mesk{\'o}, Bertalan},
  journal={Journal of medical Internet research},
  volume={25},
  pages={e50638},
  year={2023},
  publisher={JMIR Publications Toronto, Canada}
}

@inproceedings{hap,
  title={Hierarchical attention propagation for healthcare representation learning},
  author={Zhang, Muhan and King, Christopher R and Avidan, Michael and Chen, Yixin},
  booktitle={Proceedings of the 26th ACM SIGKDD International Conference on Knowledge Discovery \& Data Mining},
  pages={249--256},
  year={2020}
}

@article{healthllm7,
  title={Prompt engineering paradigms for medical applications: scoping review},
  author={Zaghir, Jamil and Naguib, Marco and Bjelogrlic, Mina and N{\'e}v{\'e}ol, Aur{\'e}lie and Tannier, Xavier and Lovis, Christian},
  journal={Journal of Medical Internet Research},
  volume={26},
  pages={e60501},
  year={2024},
  publisher={JMIR Publications Toronto, Canada}
}

@article{gpt,
  title={Gpt (generative pre-trained transformer)—A comprehensive review on enabling technologies, potential applications, emerging challenges, and future directions},
  author={Yenduri, Gokul and Ramalingam, M. and Chemmalar Selvi, G. and Supriya, Y. and Srivastava, Gautam and Maddikunta, Praveen Kumar Reddy and Deepti Raj, G. and Jhaveri, Rutvij H. and Prabadevi, B. and Wang, Weizheng and Vasilakos, Athanasios V. and Gadekallu, Thippa Reddy},
  journal={IEEE access},
  volume={12},
  pages={54608--54649},
  year={2024},
  publisher={IEEE}
}

@article{hu2022lora,
  title={Lora: Low-rank adaptation of large language models},
  author={Hu, Edward J and Shen, Yelong and Wallis, Phillip and Allen-Zhu, Zeyuan and Li, Yuanzhi and Wang, Shean and Wang, Lu and Chen, Weizhu},
  journal={arXiv preprint arXiv:2106.09685},
  year={2021}
}

@inproceedings{zhang2021grasp,
  title={GRASP: generic framework for health status representation learning based on incorporating knowledge from similar patients},
  author={Zhang, Chaohe and Gao, Xin and Ma, Liantao and Wang, Yasha and Wang, Jiangtao and Tang, Wen},
  booktitle={Proceedings of the AAAI conference on artificial intelligence},
  volume={35},
  pages={715--723},
  year={2021}
}

@inproceedings{ma2020adacare,
  title={Adacare: Explainable clinical health status representation learning via scale-adaptive feature extraction and recalibration},
  author={Ma, Liantao and Gao, Junyi and Wang, Yasha and Zhang, Chaohe and Wang, Jiangtao and Ruan, Wenjie and Tang, Wen and Gao, Xin and Ma, Xinyu},
  booktitle={Proceedings of the AAAI Conference on Artificial Intelligence},
  volume={34},
  pages={825--832},
  year={2020}
}

@inproceedings{gao2020stagenet,
  title={Stagenet: Stage-aware neural networks for health risk prediction},
  author={Gao, Junyi and Xiao, Cao and Wang, Yasha and Tang, Wen and Glass, Lucas M and Sun, Jimeng},
  booktitle={Proceedings of the web conference 2020},
  pages={530--540},
  year={2020}
}

@article{zhu2024prompting,
  title={Prompting large language models for zero-shot clinical prediction with structured longitudinal electronic health record data},
  author={Zhu, Yinghao and Wang, Zixiang and Gao, Junyi and Tong, Yuning and An, Jingkun and Liao, Weibin and Harrison, Ewen M and Ma, Liantao and Pan, Chengwei},
  journal={arXiv preprint arXiv:2402.01713},
  year={2024}
}

@inproceedings{xu-etal-2024-ram,
  title = {RAM-EHR: Retrieval Augmentation Meets Clinical Predictions on Electronic Health Records},
  author = {Xu, Ran  and Shi, Wenqi  and Yu, Yue  and Zhuang, Yuchen  and Jin, Bowen  and Wang, May Dongmei  and Ho, Joyce  and Yang, Carl},
  editor = {Ku, Lun-Wei  and Martins, Andre  and Srikumar, Vivek},
  booktitle = {Proceedings of the 62nd Annual Meeting of the Association for Computational Linguistics (Volume 2: Short Papers)},
  month = aug,
  year = "2024",
  address = "Bangkok, Thailand",
  publisher = "Association for Computational Linguistics",
  pages = "754--765",
}

@article{wolf2019huggingface,
  title={Huggingface's transformers: State-of-the-art natural language processing},
  author={Wolf, Thomas and Debut, Lysandre and Sanh, Victor and Chaumond, Julien and Delangue, Clement and Moi, Anthony and Cistac, Pierric and Rault, Tim and Louf, R{\'e}mi and Funtowicz, Morgan and others},
  journal={arXiv preprint arXiv:1910.03771},
  year={2019}
}

@article{cheong2023adaptive,
  title={Adaptive integration of categorical and multi-relational ontologies with ehr data for medical concept embedding},
  author={Cheong, Chin Wang and Yin, Kejing and Cheung, William K and Fung, Benjamin CM and Poon, Jonathan},
  journal={ACM Transactions on Intelligent Systems and Technology},
  volume={14},
  number={6},
  pages={1--20},
  year={2023},
  publisher={ACM New York, NY}
}

@article{ouyang2022training,
  title={Training language models to follow instructions with human feedback},
  author={Ouyang, Long and Wu, Jeffrey and Jiang, Xu and Almeida, Diogo and Wainwright, Carroll and Mishkin, Pamela and Zhang, Chong and Agarwal, Sandhini and Slama, Katarina and Ray, Alex and others},
  journal={Advances in neural information processing systems},
  volume={35},
  pages={27730--27744},
  year={2022}
}

@article{rafailov2023direct,
  title={Direct preference optimization: Your language model is secretly a reward model},
  author={Rafailov, Rafael and Sharma, Archit and Mitchell, Eric and Manning, Christopher D and Ermon, Stefano and Finn, Chelsea},
  journal={Advances in neural information processing systems},
  volume={36},
  pages={53728--53741},
  year={2023}
}

@inproceedings{reprompt,
  title={RePrompT: Recurrent Prompt Tuning for Integrating Structured EHR Encoders with Large Language Models},
  author={H Moghaddam, Arya and Ross, Drew and Kerdabadi, Mohsen Nayebi and Wang, Dongjie and Yao, Zijun},
  booktitle={Findings of the Association for Computational Linguistics: ACL 2026},
  pages={26766--26778},
  year={2026}
}

@inproceedings{wu2025rlpf,
  title={Rlpf: Reinforcement learning from prediction feedback for user summarization with llms},
  author={Wu, Jiaxing and Ning, Lin and Liu, Luyang and Lee, Harrison and Wu, Neo and Wang, Chao and Prakash, Sushant and O'Banion, Shawn and Green, Bradley and Xie, Jun},
  booktitle={Proceedings of the AAAI Conference on Artificial Intelligence},
  volume={39},
  pages={25488--25496},
  year={2025}
}

@article{schulman2017proximal,
  title={Proximal Policy Optimization Algorithms},
  author={Schulman, John and Wolski, Filip and Dhariwal, Prafulla and Radford, Alec and Klimov, Oleg},
  journal={arXiv preprint arXiv:1707.06347},
  year={2017}
}

@article{zheng2023secrets,
  title={Secrets of rlhf in large language models part i: Ppo},
  author={Zheng, Rui and Dou, Shihan and Gao, Songyang and Hua, Yuan and Shen, Wei and Wang, Binghai and Liu, Yan and Jin, Senjie and Liu, Qin and Zhou, Yuhao and others},
  journal={arXiv preprint arXiv:2307.04964},
  year={2023}
}

@article{liu2022note,
  title={“Note Bloat” impacts deep learning-based NLP models for clinical prediction tasks},
  author={Liu, Jinghui and Capurro, Daniel and Nguyen, Anthony and Verspoor, Karin},
  journal={Journal of biomedical informatics},
  volume={133},
  pages={104149},
  year={2022},
  publisher={Elsevier}
}

@inproceedings{kerdabadi2025multi,
  title={Multi-ontology integration with dual-axis propagation for medical concept representation},
  author={Nayebi Kerdabadi, Mohsen and H Moghaddam, Arya and Wang, Dongjie and Yao, Zijun},
  booktitle={Proceedings of the 34th ACM International Conference on Information and Knowledge Management},
  pages={2201--2211},
  year={2025}
}

@article{jiang2023health,
  title={Health system-scale language models are all-purpose prediction engines},
  author={Jiang, Lavender Yao and Liu, Xujin Chris and Nejatian, Nima Pour and Nasir-Moin, Mustafa and Wang, Duo and Abidin, Anas and Eaton, Kevin and Riina, Howard Antony and Laufer, Ilya and Punjabi, Paawan and others},
  journal={Nature},
  volume={619},
  number={7969},
  pages={357--362},
  year={2023},
  publisher={Nature Publishing Group UK London}
}

@article{huang2019clinicalbert,
  title={Clinicalbert: Modeling clinical notes and predicting hospital readmission},
  author={Huang, Kexin and Altosaar, Jaan and Ranganath, Rajesh},
  journal={arXiv preprint arXiv:1904.05342},
  year={2019}
}

@inproceedings{yadav2021reinforcement,
  title={Reinforcement learning for abstractive question summarization with question-aware semantic rewards},
  author={Yadav, Shweta and Gupta, Deepak and Abacha, Asma Ben and Demner-Fushman, Dina},
  booktitle={Proceedings of the 59th Annual Meeting of the Association for Computational Linguistics and the 11th International Joint Conference on Natural Language Processing (Volume 2: Short Papers)},
  pages={249--255},
  year={2021}
}

@inproceedings{yoon2025aspect,
  title={Aspect-oriented summarization for psychiatric short-term readmission prediction},
  author={Yoon, WonJin and Ren, Boyu and Thomas, Spencer and Kim, Chanhwi and Savova, Guergana K and Hall, Mei-Hua and Miller, Timothy A},
  booktitle={Proceedings of the 2025 Conference on Empirical Methods in Natural Language Processing},
  pages={28037--28054},
  year={2025}
}

@article{tayefi2021challenges,
  title={Challenges and opportunities beyond structured data in analysis of electronic health records},
  author={Tayefi, Maryam and Ngo, Phuong and Chomutare, Taridzo and Dalianis, Hercules and Salvi, Elisa and Budrionis, Andrius and Godtliebsen, Fred},
  journal={Wiley Interdisciplinary Reviews: Computational Statistics},
  volume={13},
  number={6},
  pages={e1549},
  year={2021},
  publisher={Wiley Online Library}
}
\clearpage
\appendix
\section{Appendix}
\subsection{RETAIN Encoder}
\label{app:retain}

For the structured EHR encoder, we use RETAIN \cite{retain}, which maps
longitudinal medical code sequences into dense patient representations through
a two-level attention mechanism: Visit-level and variable-level attention. 
The visit-level attention captures the relative
importance of the visit embeddings $\{\boldsymbol{v}_{i,j}\}_{j=1}^{t}$:
\begin{equation}
\label{eq:att_1}
\{\boldsymbol{q}_{i,j}\}_{j=1}^{T_i} =
\mathrm{GRU}_{\alpha}(\{\boldsymbol{v}_{i,j}\}_{j=1}^{T_i})
\end{equation}

\begin{equation}
\label{eq:softmax}
\{\bm{\alpha}_{i,j}\}_{j=1}^{t} =
\mathrm{Softmax}(\boldsymbol{w}_{\alpha} \{\boldsymbol{q}_{i,j}\}_{j=1}^{t})
\end{equation}

The variable-level attention assigns feature-wise weights to
the visit representation:
\begin{equation}
\label{eq:att_2}
\{\boldsymbol{z}_{i,j}\}_{j=1}^{t} =
\mathrm{GRU}_{\beta}(\{\boldsymbol{v}_{i,j}\}_{j=1}^{t})
\end{equation}

\begin{equation}
\label{eq:tanh}
\{\bm{\beta}_{i,j}\}_{j=1}^{t} =
\tanh(\boldsymbol{w}_{\beta}\{\boldsymbol{z}_{i,j}\}_{j=1}^{t})
\end{equation}

The final patient representation is computed by combining the two attention-weighted signals:
\begin{equation}
\label{eq:attention_dots}
\boldsymbol{k}_{i,t} =
\sum_{j=1}^{t} \bm{\alpha}_{i,j} \bm{\beta}_{i,j} \odot \boldsymbol{v}_{i,j}
\end{equation}

\begin{equation}
\label{eq:enc}
\boldsymbol{h}_{i,t} =
\boldsymbol{w}_{\mathrm{enc}} \boldsymbol{k}_{i,t} + b_{\mathrm{enc}}
\end{equation}

The model outputs a sequence of visit-level embeddings $\boldsymbol{h}_{i,t}$, where the final embedding at visit $T_i$ is used as the patient-level EHR representation, denoted by $\boldsymbol{h}^{\text{ehr}}_i = \boldsymbol{h}_{i,T_i}$.

\subsection{Baselines}
\label{app:baselines}
In this research, we compare RASPER with both established healthcare deep learning methods and LLM-based approaches. The deep learning baselines include \textbf{Deepr} \cite{nguyen2016mathtt}, which represents patient records as sequences of coded events with time gaps and hospital transfers and applies a CNN for prediction; \textbf{RETAIN} \cite{retain}, which uses a dual-RNN attention mechanism to capture the interpretable influence of visits and medical features;  
\textbf{GRASP} \cite{zhang2021grasp}, which improves EMR-based prediction by identifying clinically similar patients; \textbf{AdaCare} \cite{ma2020adacare}, which captures both short- and long-term biomarker variations and adaptively emphasizes patient-specific risk factors; \textbf{StageNet} \cite{gao2020stagenet}, a stage-aware neural network that models disease progression using LSTM and stage-adaptive convolution; \textbf{ADORE} \cite{cheong2023adaptive}, which uses attention to adapt medical ontology category embeddings to EHR data for clinical prediction; and \textbf{ARCI} \cite{ARCI}, which disentangles coexisting temporal medical intents across sequential visits. We also compare RASPER with three recent LLM-based healthcare prediction methods: \textbf{GraphCare} \cite{jiang2023graphcare}, which extracts knowledge from LLMs and external biomedical knowledge graphs to construct patient-specific KGs and train a Bi-attention Augmented GNN for clinical prediction; \textbf{LINKO} \cite{kerdabadi2025multi}, which uses LLM-augmented initialization and dual-axis knowledge propagation to jointly learn medical concept representations across multiple ontology graphs; and \textbf{RePrompT} \cite{reprompt}, which integrates structured EHR encoders with LLMs through time-aware prompt tuning, preserving longitudinal visit information and injecting cohort-level representations via trainable prompt tokens.



\subsection{Inference Time Comparison}

\begin{table}[ht]
\centering
\small
\caption{Inference time comparison for a batch of 12 patients.}
\label{tab:inference_time}
\begin{tabular}{lc}
\toprule
\textbf{Model} & \textbf{Time (s)} \\
\midrule
Deepr & 0.09 \\
RETAIN & 0.06 \\
GRASP & 0.17 \\
AdaCare & 0.05 \\
StageNet & 0.11 \\
Adore & 0.21 \\
ARCI & 0.92 \\
GraphCare & 0.34 \\
LINKO & 0.24 \\
RePrompT & 1.38 \\

RASPER & 1.87 \\
\bottomrule
\end{tabular}
\end{table}

In Table \ref{tab:inference_time}, we compare the proposed method with existing baselines to provide an inference time analysis. While our model is more complex than RePrompT, the observed performance gains suggest that this additional complexity is justified and contributes meaningful predictive benefit.

\end{document}